\documentclass[11pt,a4paper]{article}
\usepackage[hyperref]{eacl2021}
\usepackage{url}
\hypersetup{breaklinks=true}
\usepackage{times}
\usepackage{latexsym}
\usepackage{amsmath}
\usepackage{amsfonts}
\usepackage{nicefrac}
\usepackage{xfrac}
\usepackage{color}
\usepackage{amsthm}
\usepackage{xspace}
\usepackage{adjustbox}
\usepackage{booktabs,makecell}
\usepackage{soul}
\usepackage{breakurl}

\usepackage{cleveref}
\crefname{section}{\S}{\S\S}
\Crefname{section}{\S}{\S\S}
\crefname{table}{Tab.}{Tables}
\crefname{figure}{Fig.}{Figures}
\crefname{algorithm}{Algorithm}{}
\crefname{equation}{Eq.}{Eq.}
\crefname{appendix}{App.}{}
\crefformat{section}{\S#2#1#3}

\usepackage[disable]{todonotes}
\makeatletter
\newcommand*\iftodonotes{\if@todonotes@disabled\expandafter\@secondoftwo\else\expandafter\@firstoftwo\fi}  
\makeatother

\usepackage{microtype}

\aclfinalcopy 

\newcommand{\xx}{\mathbf{x}}
\newcommand{\yy}{\mathbf{y}}

\newcommand{\calV}{\mathcal{V}}

\newcommand{\vtheta}{{\boldsymbol \theta}}
\newcommand{\ptheta}{p_{\vtheta}}
\newcommand{\eos}{\textsc{eos}\xspace}
\newcommand{\bos}{\textsc{bos}\xspace}
\newcommand{\ethz}{1}
\newcommand{\ucambridge}{2}
\newcommand{\defn}[1]{\textbf{#1}}

\DeclareMathOperator*{\argmax}{argmax}

\title{Searching for Search Errors in Neural Morphological Inflection}

\author{Martina Forster$^\ethz$~\;~Clara Meister$^\ethz$~\;~Ryan Cotterell$^{\ethz, \ucambridge}$ \\
  $^\ethz$ETH Z\"{u}rich~\;~$^\ucambridge$University of Cambridge \\
  \texttt{martfors@ethz.ch}~\;~\texttt{\{clara.meister, ryan.cotterell\}@inf.ethz.ch}}
\date{}

\begin{document}
\maketitle
\begin{abstract}
    Neural sequence-to-sequence models are currently the predominant choice for language generation tasks. Yet, on word-level tasks, exact inference of these models reveals the empty string is often the global optimum. Prior works have speculated this phenomenon is a result of the inadequacy of neural models for language generation. However, in the case of morphological inflection, we find that the empty string is almost never the most probable solution under the model. Further, greedy search often finds the global optimum. These observations suggest that the poor calibration of many neural models may stem from characteristics of a specific subset of tasks rather than general ill-suitedness of such models for language generation. \looseness=-1

\end{abstract}

\section{Introduction}
\label{introduction}
Neural sequence-to-sequence models are omnipresent in the field of natural language processing due to their impressive performance. 
They hold state of the art on a myriad of tasks, e.g., neural machine translation \cite[NMT;][]{ott-etal-2018-scaling} and abstractive summarization \cite[AS;][]{lewis2019bart}. 
Yet, an undesirable property of these models has been repeatedly observed in word-level tasks: When using beam search as the decoding strategy, increasing the beam width beyond a size of $k=5$ often leads to a drop in the  quality of solutions \cite{murray-chiang-2018-correcting, yang-etal-2018-breaking,pmlr-v97-cohen19a}. 
Further, in the context of NMT, it has been shown that the empty string is frequently the most-probable solution under the model \cite{stahlberg-byrne-2019-nmt}. Some suggest this is a manifestation of the general inadequacy of neural models for language generation tasks \cite{koehn-knowles-2017-six, kumar2019calibration,holtzman2019curious,stahlberg_phd}; in this work, we find evidence demonstrating otherwise.\looseness=-1  

\begin{table}

    \centering
    \adjustbox{width=\linewidth}{
    \begin{tabular}{@{}c|llll@{}} 
        \toprule
        & $k=1$ & $k=10$ &$k=100$ & $k=500$ \\
        \hline
        \bf NMT & 63.1\% & 46.1\% & 44.3\% & 6.4\% \rule{0pt}{3ex}\\ 
        \bf MI  & ~0.8\% & ~0.0\% & ~0.0\% & 0.0\% \\ 
        \bottomrule
    \end{tabular}}
    \caption{Percentage of search errors---which we define as instances where the search strategy does not find the global optimum under the model---for Transformers trained on IWSLT'14 De-En (NMT) and SIGMORPHON 2020 (Morphological Inflection; MI) when decoding with beam search for varying beam widths ($k$). MI results are averaged across  languages.}
    \label{tab:search_errors_nmt}
\end{table}

Sequence-to-sequence transducers for character-level tasks often follow the architectures of their word-level counterparts \cite{faruqui-etal-2016-morphological, lee-etal-2017-fully}, and have likewise achieved state-of-the-art performance on e.g., morphological inflection generation \cite{wu2020applying} and grapheme-to-phoneme conversion \cite{G2P}. 
Given prior findings, we might expect to see the same degenerate behavior in these models---however, we do not. 
We run a series of experiments on morphological inflection (MI) generators to explore whether neural transducers for this task are similarly poorly calibrated, i.e. are far from the true distribution $p(\yy\mid\xx)$. 
We evaluate the performance of two character-level sequence-to-sequence transducers using different decoding strategies; our results, previewed in \cref{tab:search_errors_nmt}, show that evaluation metrics do not degrade with larger beam sizes as in NMT or AS. 
Additionally, only in extreme circumstances, e.g., low-resource settings with less than 100 training samples, is the empty string ever the global optimum under the model.\looseness=-1

Our findings directly refute the claim that neural architectures are inherently inadequate for modeling language generation tasks. Instead, our results admit two potential causes of the degenerate behavior observed in tasks such as NMT and AS: (1) lack of a deterministic mapping between input and output and (2) a (perhaps irreparable) discrepancy between sample complexity and training resources. Our results alone are not sufficient to accept or reject either hypothesis, and thus we leave these as future research directions.\looseness=-1

\section{Neural Transducers}
Sequence-to-sequence transduction is the transformation of an input sequence into an output sequence. Tasks involving this type of transformation are often framed probabilistically, i.e., we model the \emph{probability} of mapping one sequence to another. On many tasks of this nature, neural sequence-to-sequence models \cite{sutskever_rnn, Bahdanau} hold state of the art.  

Formally, a neural sequence-to-sequence model defines a probability distribution $\ptheta(\yy\!\mid\!\xx)$ parameterized by a neural network with a set of learned weights $\vtheta$ for an input sequence $\xx = \langle x_1, x_2, \dots\rangle$ and output sequence $\yy = \langle y_1, y_2, \dots \rangle$. 
Morphological inflection and NMT are two such tasks, wherein our outputs are both strings.
Neural sequence-to-sequence models are typically locally normalized, i.e. $\ptheta$ factorizes as follows:\looseness=-1
\begin{equation}\label{eq:sequence_probability}
    \ptheta(\yy \!\mid \!\xx) = \prod_{t=1}^{|\yy|}p_\vtheta(y_t \mid\! \xx, \yy_{<t})
\end{equation}
\noindent Given a vocabulary $\mathcal{V}$, each conditional $\ptheta$ is a distribution over $\mathcal{V} \cup \{\eos\}$ and
$y_0 := \bos$. 
We consider $\ptheta(\yy \!\mid\! \xx)$ to be \defn{well-calibrated} if its probability estimates are representative of the true likelihood that a solution $\yy$ is correct.

\paragraph{Morphological Inflection.}
In the task of morphological inflection, $\xx$ is an encoding of the lemma concatenated with a flattened morphosyntactic description (MSD) and $\yy$ is the target inflection. 
As a concrete example, consider inflecting the German word \textit{Bruder} into the genitive plural, as shown in \cref{tab:inflection_example}. Then, $\xx$ is the string $\langle \texttt{B r u d e r GEN PL}\rangle$
and $\yy$ is the string $\langle \texttt{B r {\"u} d e r}\rangle$.
As this demonstrates, morphological inflection generation is, by its nature, modeled at the character level \cite{faruqui-etal-2016-morphological,wu-cotterell-2019-exact}, i.e., our target vocabulary $\calV$ is a set of characters in the language.
Note that $\yy \in \calV^*$, but $\xx \not\in \calV^*$ due to the additional encoding of the MSD.
This stands in contrast to NMT, which is typically performed on a (sub)word level, making the vocabulary size orders of magnitude larger.\looseness=-1 

Another important differentiating factor of morphological inflection generation in comparison to many other generation tasks in NLP is the one-to-one mapping between source and target.\footnote{While there are cases where there exist multiple inflected forms of a lemma, e.g., in English the past tense of \textit{dream} can be realized as either \textit{dreamed} or \textit{dreamt}, these cases (termed ``overabundance'') are rare \cite{OverabundanceinMorphology}.} 
In contrast, there are almost always many correct ways to translate a sentence into another language or to summarize a large piece of text;
this characteristic manifests itself in training data where a single phrase has instances of different mappings, making tasks such as translation and summarization inherently ambiguous. 

\begin{table}
\centering
\small
\adjustbox{max width=\linewidth}{

    \begin{tabular}{@{}l|ll@{}} 
        \toprule
        & Singular & Plural  \\
        \hline
        Nominativ & Bruder & Br\"uder \rule{0pt}{3ex} \\ 
        Genitiv & Bruders & Br\"uder  \\ 
        Dativ & Bruder & Br\"udern \\ 
        Akkusativ & Bruder & Br\"uder \\ 
        \bottomrule
    \end{tabular}}
    \caption{Inflection table for the German word \textit{Bruder}}
    \label{tab:inflection_example}
\end{table}

\section{Decoding}
In the case of probabilistic models, the decoding problem is the search for the most-probable sequence among valid sequences $\calV^*$ under the model $\ptheta$:\looseness=-1
\begin{equation}\label{eq:decoding}
    \yy^\star = \argmax_{\yy \in \calV^*} \log \ptheta(\yy \mid \xx)
\end{equation}
\noindent This problem is also known as maximum-a-posteriori (MAP) inference. 
Decoding is often performed with a heuristic search method such as greedy or beam search 
\cite{reddy-1977}, since performing exact search can be computationally expensive, if not impossible.\footnote{The search space is exponential in the sequence length and due to the non-Markov nature of (typical) neural transducers, dynamic-programming techniques are not helpful.} 
While for a deterministic task, greedy search is optimal under a Bayes optimal model,\footnote{Under such a model, the correct token $y_i$ at time step $i$ will be assigned all probability mass.} most text generation tasks benefit from using beam search. 
However, text quality almost invariably decreases for beam sizes larger than $k=5$. 
This phenomenon is sometimes referred to as the \defn{beam search curse}, and has been investigated in detail by a number of scholarly works \cite{koehn-knowles-2017-six, murray-chiang-2018-correcting, yang-etal-2018-breaking, stahlberg-byrne-2019-nmt, pmlr-v97-cohen19a, eikema2020map}.
\begin{table*}
\centering 
\adjustbox{max width=\textwidth}{
\begin{tabular}{ @{}l|llll|llll@{}} 
\toprule
  & \multicolumn{4}{c|}{\bf Transformer} & \multicolumn{4}{c}{\bf HMM}\\
& $k=1$ & $k=10$ & $k=100$ & Dijkstra & $k=1$ & $k=10$ & $k=100$ & Dijkstra\\ \hline
Overall & 90.34\% & 90.37\% & 90.37\% & 90.37\% & 86.03\% & 85.62\% & 85.60\% & 85.60\% \rule{0pt}{3ex}\\ 
Low-resource & 84.10\% & 84.12\% & 84.12\% & 84.12\% & 70.99\% & 69.37\% & 69.31\% & 69.31\%\\ 
High-resource & 94.05\% & 94.08\% & 94.08\% & 94.08\% & 93.60\% & 93.72\% & 93.72\% & 93.72\%\\ 
\bottomrule
\end{tabular}}
\caption{Prediction accuracy (averaged across languages) by decoding strategy for Transformer and HMM. We include breakdown for low-resource and high-resource trained models. $k$ indicates beam width.}
\label{tab:accuracies} 
\end{table*}

Exact decoding can be seen as the case of beam search where the beam size is effectively stretched to infinity.\footnote{This interpretation is useful when comparing with beam search with increasing beam widths.} By considering the complete search space, it finds the globally best solution under the model $\ptheta$. While, as previously mentioned, 
exact search can be computationally expensive, we can employ efficient search strategies due to some properties of $\ptheta$. Specifically, from \cref{eq:sequence_probability}, we can see that the scoring function for sequences $\yy$ is monotonically decreasing in $t$. We can therefore find the provably optimal solution with Dijkstra's algorithm \cite{dijkstra1959note}, which terminates and returns the global optimum the first time it encounters an \eos. Additionally, to prevent a large memory footprint, we can lower-bound the search using any complete hypothesis, e.g., the empty string or a solution found by beam search \cite{stahlberg-byrne-2019-nmt, meister+al.tacl20}. That is, we can prematurely stop exploring solutions whose scores become less than these hypotheses at any point in time. Although exact search is an exponential-time method in this setting, we see that, in practice, it terminates quickly due to the peakiness of $\ptheta$ (see \cref{sec:time}). 
While the effects of exact decoding and beam search decoding with large beam widths have been explored for a number of word-level tasks \cite{stahlberg-byrne-2019-nmt, pmlr-v97-cohen19a, eikema2020map}, to the best of our knowledge, they have not yet been explored for any character-level sequence-to-sequence tasks.\looseness=-1

\section{Experiments}

We run a series of experiments using different decoding strategies to generate predictions from morphological inflection generators. 
We report results for two near-state-of-the-art models: a multilingual Transformer \cite{wu2020applying} and a (neuralized) hidden Markov model \cite[HMM;][]{wu-cotterell-2019-exact}. 
For reproducibility, we mimic their proposed architectures and exactly follow their  data pre-processing steps, training strategies and hyperparameter settings.\footnote{  \url{https://github.com/shijie-wu/neural-transducer/tree/sharedtasks}}

\paragraph{Data.}
We use the data provided by the SIGMORPHON 2020 shared task \cite{vylomova2020sigmorphon}, which features lemmas, inflections, and corresponding MSDs in the UniMorph schema \cite{kirov-etal-2018-unimorph} in 90 languages in total. 
The set of languages is typologically diverse (spanning 18 language families) and contains both high- and low-resource examples, providing a spectrum over which we can evaluate model performance. 
The full dataset statistics can be found on the task homepage.\footnote{\url{https://sigmorphon.github.io/sharedtasks/2020/task0/}} When reporting results, we consider languages with $< 1000$ and $\geq 10000$ training samples as low- and high-resource, respectively.

\paragraph{Decoding Strategies.}
We decode morphological inflection generators using exact search and beam search for a range of beam widths. 
We use the SGNMT library for decoding \cite{stahlberg2017sgnmt} albeit adding Dijkstra's algorithm.\looseness=-1

\begin{table}
\centering 
\adjustbox{max width=\linewidth}{
\begin{tabular}{ @{}l|llcl@{}} 
\toprule
   & \bf \makecell[c]{Beam\\$k\!=\!1$} & \bf \makecell[c]{Beam\\$k\!=\!10$} & \bf Optimum &\bf \makecell[c]{Empty\\String}\\ \hline
Transformer  &  -0.619 & -0.617 & -0.617 & ~~-6.56 \rule{0pt}{3ex}\\
HMM  & -1.08 & -0.89 & -0.80~~ &  -20.15\\ 
\bottomrule
\end{tabular}}
\caption{Average log probability of inflections generated with various decoding strategies and the empty string (averaged across all languages).}
\label{tab:empty_prob_vs_log_prob} 
\end{table}

\subsection{Results}

\begin{figure}
    \centering
    \includegraphics[width=0.49\textwidth]{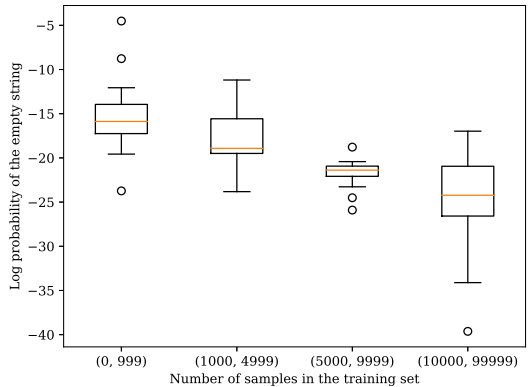}
    \caption{Average (log) probability of the empty string for different training dataset sizes for HMM.}
    \label{fig:empty_string_prob_vs_trn_size_mono-hmm}
\end{figure}

\cref{tab:accuracies} shows that the accuracy of predictions from neural MI generators generally does not decrease when larger beam sizes are used for decoding; this observation holds for both model architectures. While it may be expected that models for low-resource languages generally perform worse than those for high-resource ones, this disparity is only  prominent for HMMs, where the difference between high- and low-resource accuracy is $\approx 24\%$ vs. $\approx 10\%$ for the Transformers. Notably, for the HMM, the global optimum under the model is the empty string far more often for low-resource languages than it is for high-resource ones (see  \cref{tab:avg_percentage_empty_strings_high_low_resource_mono-hmm_dijkstra}). We can explicitly see the inverse relationship between the log-probability of the empty string and resource size in \cref{fig:empty_string_prob_vs_trn_size_mono-hmm}. 
In general, across models for all 90 languages, the global optimum is rarely the empty string (\cref{tab:avg_percentage_empty_strings_high_low_resource_mono-hmm_dijkstra}). Indeed, under the Transformer-based transducer, the empty string was \emph{never} the global optimum. This is in contrast to the findings of \citet{stahlberg-byrne-2019-nmt}, who found for word-level NMT that the empty string was the optimal translation in more than 50\% of cases, even under state-of-the-art models. Rather, the average log-probabilities of the empty string (which is quite low) and the chosen inflection lie far apart (\cref{tab:empty_prob_vs_log_prob}).\looseness=-1

\section{Discussion}
Our findings admit two potential hypotheses for poor calibration of neural models in certain language generation tasks, a phenomenon we do not observe in morphological inflection. First, the tasks in which we observe this property are ones that lack a deterministic mapping, i.e. tasks for which there may be more than one correct solution for any given input. As a consequence, probability mass may be spread over an arbitrarily large number of hypotheses \cite{ott2018analyzing,eikema2020map}. In contrast, 
the task of morphological inflection has a near-deterministic mapping. We observe this empirically in \cref{tab:empty_prob_vs_log_prob}, which shows that the probability of the global optimum on average covers most of the available probability mass---a phenomenon also observed by \citet{peters-martins-2019-ist}. Further, as shown in \cref{tab:search_errors}, the dearth of search errors even when using greedy search suggests there are rarely competing solutions under the model. We posit it is the lack of ambiguity in morphological inflection that allows for the well-calibrated models we observe.\looseness=-1

Second, our experiments contrasting high- and low-resource settings indicate insufficient training data 
may be the main cause of the poor calibration in sequence-to-sequence models for language generation tasks. We observe that models for MI trained on fewer data typically place more probability mass on the empty string. As an extreme example, we consider the case of the Zarma language, whose training set consists of only 56 samples. Under the HMM, the average log-probability of the generated inflection and empty string are very close ($-8.58$ and $-8.77$, respectively). Furthermore, on the test set, the global optimum of the HMM model for Zarma is the empty string 81.25\% of the time.\looseness=-1
\begin{table}
\begin{center}
\begin{tabular}{ @{}l|cc@{} } 
 \toprule
  & \textbf{HMM} & \textbf{Transformer}\\  \hline 
Overall & ~~ 2.03\%~~~~ & 0\%\rule{0pt}{3ex}\\
Low-resource & ~~ 8.65\% ~~~ & 0\%\\ 
High-resource & ~~ 0.0002\% & 0\% \\ 
 \bottomrule
\end{tabular}
\end{center}
\caption{Average percentage of empty strings when decoding with exact inference for HMM and Transformer, with resource group breakdown.}
\label{tab:avg_percentage_empty_strings_high_low_resource_mono-hmm_dijkstra}
\end{table}

\begin{table}
\small
    \centering
    \adjustbox{max width=\linewidth}{
    \begin{tabular}{@{}l|llll@{}} 
        \toprule
        & $k = 1$ & $k = 10$ & $k = 100$ & $k = 200$ \\
        \hline
        HMM & 6.20\% & 2.33\% & 0.001\% & 0.0\% \rule{0pt}{3ex}\\ 
        Transformer & 0.68\% & 0.0\% & 0.0\% & 0.0\% \\ 
        \bottomrule
    \end{tabular}}
    \caption{Average percentage of search errors (averaged across languages) for beam search with beam width $k$.\looseness=-1}
    \label{tab:search_errors}
\end{table}
From this example, we can conjecture that lack of sufficient training data may manifest itself as the (relatively) high probability of the empty string or the (relatively) low probability of the optimum.
We can extrapolate to models for NMT and other word-level tasks, for which we frequently see the above phenomenon. Specifically, our experiments suggest that when neural language generators frequently place high probability on the empty string, there may be a discrepancy between the available training resources and the number of samples needed to successfully learn the target function. While this at first seems an easy problem to fix, we  expect the number of resources needed in tasks such as NMT and AS is much larger than that for MI if not due to the size of the output space alone; perhaps so large that they are essentially unattainable. Under this explanation, for certain tasks, there may not be a straightforward fix to the degenerate behavior observed in some neural language generators.\looseness=-1

\section{Conclusion}
In this work, we
investigate whether the poor calibration often seen in sequence-to-sequence models for word-level tasks also occurs in models for morphological inflection.
We find that character-level models for morphological inflection are generally well-calibrated, i.e. the probability of the globally best solution is almost invariably much higher than that of the empty string. 
This suggests the degenerate behavior observed in neural models for certain word-level tasks is not due to the inherent incompatibility of neural models for language generation. 
Rather, we find evidence that poor calibration may be linked to specific characteristics of a subset of these task, and suggest directions for future exploration of this phenomenon.

\bibliography{anthology,eacl2021}
\bibliographystyle{acl_natbib}

\clearpage

\appendix
\label{appendix}
\section{Timing}\label{sec:time}
\begin{table}[!h]
\small 
\centering 
\adjustbox{width=\linewidth}{
\begin{tabular}{ @{}l|cc|cc@{}} 
\toprule
  & \multicolumn{2}{c|}{\bf Transformer} & \multicolumn{2}{c}{\bf HMM}\\
& $k=1$  & Dijkstra & $k=1$ & Dijkstra\\ \hline
Overall & 0.082  & 0.091 & 0.016 & 0.027 \rule{0pt}{3ex}\\ 
Low-resource & 0.072  & 0.082 & 0.013 & 0.032\\ 
High-resource & 0.075  & 0.083 & 0.017 & 0.026 \\ 
\bottomrule
\end{tabular}}
\caption{Average time (s) for inflection generation by decoding strategy. Breakdown by resource group is included. }
\label{tab:time} 
\end{table}

\end{document}